\documentclass[journal]{IEEEtran}
\usepackage[numbers,sort&compress]{natbib}
\usepackage{amsmath}
\usepackage{amsfonts}
\usepackage{xcolor}
\usepackage{url}
\usepackage{hyperref}
\usepackage{ulem}
\usepackage{booktabs}

\usepackage[pdftex]{graphicx}

\title{Adherence Forecasting for Guided Internet-Delivered Cognitive Behavioral Therapy: A Minimally Data-Sensitive Approach}
\author{Ulysse Côté-Allard$^{\dagger, 1}$, Minh~H.~Pham$^{\dagger, 1}$, Alexandra K. Schultz$^{2}$, Tine~Nordgreen$^{3}$, Jim~Torresen$^{1}$ %
\thanks{}
\thanks{$\dagger$ These authors share first authorship and are listed in alphabetical order}
\thanks{$^{1}$Minh H. Pham, Ulysse Côté-Allard and Jim Torresen are with the depatment of Informatics at the University of Oslo, Oslo, Norway. Corresponding author: Minh H. Pham ({\tt\small hoangmph@uio.no}) and Ulysse Côté-Allard ({\tt\small ulysseca@uio.no})}%
\thanks{$^{2}$Alexandra K. Schultz is with the faculty of law at the University of Oslo, Oslo, Norway.}
\thanks{$^{3}$Tine Nordgreen is with the Department of Clinical Psychology at the University of Bergen, Bergen, Norway and the Division of Psychiatry, Haukeland University Hospital, Bergen, Norway.}%
}

\begin{document}

\maketitle

\begin{abstract}

Internet-delivered psychological treatments (IDPT) are seen as an effective and scalable pathway to improving the accessibility of mental healthcare. Within this context, treatment adherence is an especially pertinent challenge to address due to the reduced interaction between healthcare professionals and patients. In parallel, the increase in regulations surrounding the use of personal data, such as the General Data Protection Regulation (GDPR), makes data minimization a core consideration for real-world implementation of IDPTs. Consequently, this work proposes a Self-Attention-based deep learning approach to perform automatic adherence forecasting, while only relying on minimally sensitive login/logout-timestamp data. This approach was tested on a dataset containing 342 patients undergoing Guided Internet-delivered Cognitive Behavioral Therapy (G-ICBT) treatment. Of these 342 patients, 101 ($\sim$30\%) were considered non-adherent (dropout) based on the adherence definition used in this work (i.e. at least eight connections to the platform lasting more than a minute over 56 days). The proposed model achieved over 70\% average balanced accuracy, after only 20 out of the 56 days ($\sim$1/3) of the treatment had elapsed. This study demonstrates that automatic adherence forecasting for G-ICBT, is achievable using only minimally sensitive data, thus facilitating the implementation of such tools within real-world IDPT platforms.
\end{abstract}


\section{Introduction}
Mental illness is associated with major individual, societal and economical challenges, which currently accounts for 20\% of the burden of disease worldwide~\cite{mental_illness_burden_of_disease_20_percent}. Fortunately, effective evidence-based somatic and psychotherapeutic treatments have been developed for a wide variety of mental disorders~\cite{efficacy_psychotherapy_vs_pharmacotherapy, CBT_works_for_anxiety_disorders, combining_pharmacotherapy_and_psychotherapy_meta_analysis, meta_analysis_long_term_efficacy_of_psychotherapy_for_depression}. Cognitive Behavioral Therapy (CBT) is a popular and effective form of psychological treatment for an array of mental disorders and is considered by some to be ``the \textit{gold standard} in the psychotherapy field''~\cite{cbt_gold_standard}. However, despite clear evidence of the effectiveness of CBT~\cite{CBT_works_for_anxiety_disorders, cbt_meta_anlasys_of_meta_analysis}, an important gap still exists between the needs of patients and the presence of affordable and timely services~\cite{INTROMAT_overall_paper, mental_health_gap_between_needs_and_what_is_offered}. This gap - which mainly stems from social stigma, lack of healthcare workers and under-prioritization of mental health services~\cite{INTROMAT_overall_paper} - combined with the high economical cost associated with mental health~\cite{mental_illness_burden_of_disease_20_percent, cost_brain_disorder, cost_global_mental_health} has created a strong demand for more affordable and accessible treatments~\cite{mental_health_gap_between_needs_and_what_is_offered}. 

\begin{figure}[!htbp]
\centering
\includegraphics[width=\linewidth]{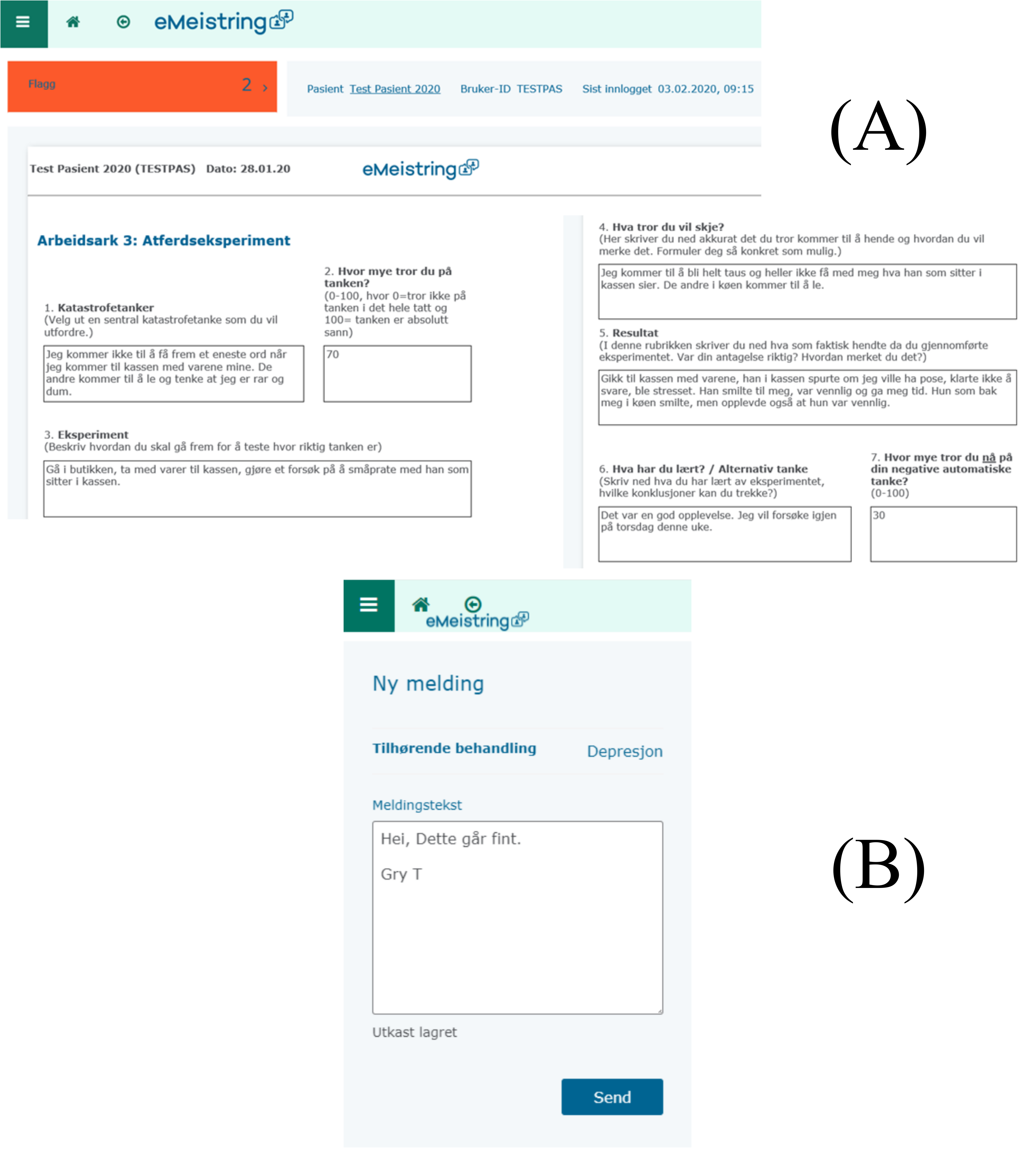}
\caption{An example of the interfaces seen on the eMeistring platform. Figure (A) showcases a homework example, while Figure (B) presents the interface cellphone interface seen when writing an email via the secure mail service within the application.}
\label{fig:emeistring_interface}
\end{figure}

Internet-delivered psychological treatments (IDPT)~\cite{internet_delivered_psychological_treatments} offer an attractive and scalable pathway to improve the accessibility of mental healthcare. In particular, Guided Internet-delivered Cognitive Behavioral Therapy (G-ICBT) has been shown to be effective for a wide range of mental illnesses and in some cases provides comparable outcomes to in-person CBT~\cite{meta_analysis_cbt_vs_g_icbt_2016, andersson2019internet}. Consequently, several platforms have been developed worldwide to provide patients access to G-ICBT. eMeistring~\cite{eMeistring_panic_disorder, eMeistring_social_anxiety_disorder} is one such platform that offers G-ICBTs for social anxiety disorder, panic disorder and depression. The three interventions available on the platform are comprised of text-based modules focusing on psychoeducation, behavioral activation, and cognitive reappraisal in the case of depression and psychoeducation, working with automatic thoughts, behavioral experiments, shifting focus, and relapse prevention for the social anxiety and panic disorders interventions. These text-based modules also feature various intervention-specific homework assignments (e.g. identifying thoughts, sleep diary registration, activity planning) to be filled within the platform itself as well as exercises (e.g. exposure to agoraphobic situations). Therapist guidance is provided at least once a week via a secure email system and when judged necessary phone calls can also be arranged. Detailed descriptions of the intervention for panic disorder~\cite{eMeistring_panic_disorder}, social anxiety disorder~\cite{eMeistring_social_anxiety_disorder} and depression~\cite{depression_eMeistring} are provided in their associated reference, while Figure~\ref{fig:emeistring_interface} showcases the platform's user interface. eMeistring is currently available to the public in parts of Norway, with plans to be deployed nationwide within the next two years.

The law of attrition~\cite{law_of_attrition_seminal_paper_2005} is a landmark paper which states that for many eHealth trials it should be expected that substantial participant dropout will occur, particularly in the case of internet-delivered treatments.
Thus, despite the effectiveness and availability of G-ICBT, patient adherence to this form of treatment remains an important challenge~\cite{adherence_guided_icbt}, as participant dropout impacts treatment outcome and results in wasted time for both the patient and clinician. 
 Unfortunately, assessing patient \textit{adherence} in the context of G-ICBT is especially challenging due to the sparsity (or lack of) direct interactions between the clinician and the patient. Further, any strategy used to assess adherence would need to require minimal resources from the clinician, as one advantage of G-ICBT is that the psychotherapist can oversee more patients. Consequently, having a tool that can automatically identify individuals likely to dropout early in the intervention based on the participant's behavior, would allow the clinician to perform meaningful, targeted intervention (e.g. providing reminders, scheduling direct interactions, modifying the treatment) more effectively.


Simultaneously, there are increasing regulations regarding the acquisition and use of personal data being implemented globally. The General Data Protection Regulation (GDPR)~\cite{GDPR}, which came into effect on May 25th 2018, is a prime example of this. Similar regulations are being adopted worldwide such as the California Consumer Privacy Act (CCPA) in the United States, the General Data Protection Law (LGPD) in Brazil and the Personal Information Protection and Electronic Documents Act (PIPEDA) in Canada. Many of these follow the blueprint of the GDPR, and this in combination with the wide territorial scope, makes the GDPR unmatched in terms of global influence~\cite{GDPR_Global_Protection}. In these regulations, there are commonly certain categories of data that evoke particular concern. For instance, article 9 of the GDPR and 11 of the LGPD place stricter requirements on the processing of data belonging to categories such as health, sexual orientation, and political opinion of the data subject. Thus, the use of such data can be cumbersome and sometimes even impossible. Generally, data generated in the context of a healthcare intervention would fall within such categorization, as it represents highly sensitive information. On top of strict requirements, one of the core tenants of the GDPR and similar regulations is the data-minimization principle (see article 5 (1)(c)). This obligates data collectors to use the minimum amount and least sensitive data necessary to achieve the purpose of the data collection. As such, being able to use the smallest amount of data which also contains the minimal amount of personal information to achieve useful adherence prediction is an auspicious target. Consequently, the main aim of this paper is to demonstrate the feasibility of minimally data-sensitive automatic adherence forecasting, solely relying on a pseudonymized user id~\cite{data_pseudonymisation_enisa_2021} and their login/logout timestamps.

Many research groups have illustrated the positive association of patient adherence to treatment outcome and have investigated how different population and treatment characteristics (e.g. demographic, psychometric, treatment credibility) relates to adherence~\cite{idpt_adherence_positivilely_impact_outcomes, adherence_leads_to_better_treatment_outcome, icbt_adherence_prediction_demographics, adherence_predictor_depression_internet_based_intervention, adherence_predictor_depression_internet_based_intervention_2, treatment_credibility_predict_dropout}. For example, Karyotaki et al.~\cite{predictors_web_based_interventions_for_depression} identified through a meta-analysis of self-guided web-based interventions for depression that gender, age, education level and co-morbid anxiety symptoms could be used as predictors of adherence. However, few works have considered the use of machine learning (ML) for adherence forecasting on an individual basis. Wallert et al.~\cite{wallert2018} trained a random forest based on information available during the first treatment (demographic, clinical, psychometric and linguistics), which achieved an accuracy of 64\% for adherence prediction of IDPT. More recently, Forsell et al.~\cite{predicting_treatment_failure_gICBT} showed the feasibility of predicting treatment failure of a 12-week G-ICBT treatment for depression, social anxiety disorder and panic disorder based on screening, pre-treatments and weekly symptom self-rating, reaching an average balanced accuracy of ~63\% after 2 weeks and ~71\% after 6 weeks. Importantly, this met the accuracy threshold of 65-70\% at which clinicians become willing to act on predictions~\cite{threshold_test_treatment_65_to_70_percent}. Both cases however rely heavily on sensitive personal medical and/or biometric data for forecasting. Further, due to the forecasting methods employed in these works, a different model has to be trained at each step of the treatment (i.e. the same model cannot be used to forecast at day 7 and day 8).

Consequently, this work's main contribution is to demonstrate the feasibility of adherence forecasting for G-ICBT using a minimally data-sensitive approach by leveraging a self-attention deep neural network~\cite{transformer}. Importantly, only a single instance of the proposed model needs to be trained to be able to forecast at any step of the treatment. The proposed method is evaluated on real-life patients using the eMeistring platform, with a collected dataset composed of 342 individuals undergoing G-ICBT treatment for depression, social anxiety disorder or panic disorder. For reproducibility purposes, the full code employed in this research is available at the following link: \href{https://github.com/UlysseCoteAllard/AdherenceForecastingG-ICBT}{github.com/UlysseCoteAllard/AdherenceForecastingG-ICBTl} alongside the dataset used.

This paper is organized as follows. The three treatment programs (depression, social anxiety disorder and panic disorder) available on the eMeistring platform alongside the data collected from these programs are described in Section~\ref{eMestring_section}. Section~\ref{adherence_prediction_section} then presents the data processing and adherence prediction methods considered in this work. Finally, the results and the associated discussion are covered in Section~\ref{results_section} and~\ref{discussion_section} respectively. 

\section{eMeistring and data collection}
\label{eMestring_section}
Since 2015, the eCoping clinic (eMeistring.no) at Haukeland University Hospital, Bergen, Norway has offered a G-ICBT for social anxiety disorder, panic disorder and depression. All patients admitted for specialized mental health treatment within the health region associated with eCoping are referred by their general practitioner. Subsequently, referred patients admitted for treatment were invited for an in-person assessment interview at the clinic. Patients were informed in the meeting about G-ICBT being one of the treatment options available. 

Patients willing to consider G-ICBT as a treatment alternative were invited to a diagnostic assessment using the Mini International Neuropsychiatric Interview (MINI)~\cite{sheehan1998mini}. Patients interested in starting G-ICBT and who fulfilled the inclusion criteria were offered G-ICBT and invited to participate in this data collection. The inclusion criteria were:

\begin{itemize}
    \item Panic disorder (with and without agoraphobia) / Social anxiety disorder / Major depressive disorder (mild and moderate) as the primary diagnosis according to the MINI
    \item 18 years old or more
    \item Not using benzodiazepines or other sedatives on a daily basis
    \item If using antidepressant, the dosage must have been stable for the preceding four weeks
    \item Able to read and write in Norwegian
    \item No current suicidal ideation
    \item No current psychosis
    \item No current substance abuse
    \item Not currently in need of other immediate treatment (i.e. due to a more severe primary diagnosis/crisis or exhibiting suicidal ideations)
    \item Has Internet access
\end{itemize}


The suicidal ideation and psychosis symptoms were assessed in a face to face clinical interview before treatment. Self-report questionnaires were further employed to evaluate suicidal ideation during treatment. “In immediate need for other treatment” meant that if the patient were in need of other treatment due to a more severe diagnosis, they were excluded. The data recording protocol was approved by The Western Regional Committee for Medical and Health Research Ethics in Norway (2015/878) and (2012/2211/REK). Written informed consent was obtained from all participants, and no financial compensation was provided. The initial dataset was comprised of 398 patients. The distribution of the participants for the considered diagnostics was as follows: panic disorder (124), social anxiety disorder (169) and depression (105). The patient characteristics are presented in Table~\ref{table:demographics}. Note that, to remove trivially non-adherent participants, only those who connected more than once to the eMeistring platform were considered in this work, resulting in a dataset containing 342 participants.

\begin{table*}[!htbp]
\centering
\caption{Patient characteristics for the three considered diagnostics.}
\begin{tabular}{ccccccccccccc}
\hline
 & \multicolumn{3}{c}{\textbf{Panic Disorder}} & \multicolumn{3}{c}{\textbf{\begin{tabular}[c]{@{}c@{}}Social Anxiety \\ Disorder\end{tabular}}} & \multicolumn{3}{c}{\textbf{Depression}} & \multicolumn{3}{c}{\textbf{Aggregated}} \\ \hline
 & n/N, Mean & \%/SD & \multicolumn{1}{c|}{Range} & n/N, Mean & \%/SD & \multicolumn{1}{c|}{Range} & n/N, Mean & \%/SD & \multicolumn{1}{c|}{Range} & n/N, Mean & \%/SD & Range \\ \cline{2-13} 
Female & 81/124 & 65.3\% & \multicolumn{1}{c|}{} & 96/169 & 56.8\% & \multicolumn{1}{c|}{} & 61/105 & 58.0\% & \multicolumn{1}{c|}{} & 238/398 & 59.80\% &  \\
Age & 35.9 & 11.8 & \multicolumn{1}{c|}{19-69} & 29.8 & 10.6 & \multicolumn{1}{c|}{18-63} & 35 & 12 & \multicolumn{1}{c|}{19-71} & 33.1 & 11.68 &  \\
Higher Education & 41/124 & 33.1\% & \multicolumn{1}{c|}{} & 49/169 & 29.0\% & \multicolumn{1}{c|}{} & 51/105 & 48.0\% & \multicolumn{1}{c|}{} & 141/398 & 35.4\% &  \\
Married/Cohabitant & 79/124 & 63.7\% & \multicolumn{1}{c|}{} & 66/166 & 39.8\% & \multicolumn{1}{c|}{} & 59/105 & 56.0\% & \multicolumn{1}{c|}{} & 204/395 & 51.7\% &  \\
Has Children & 66/124 & 53.2\% & \multicolumn{1}{c|}{} & 41/167 & 24.6\% & \multicolumn{1}{c|}{} & 42/105 & 40.0\% & \multicolumn{1}{c|}{} & 149/396 & 40.4\% &  \\
Years with complaints & 8.5 & 8.8 & \multicolumn{1}{c|}{0-36} & 13.8 & 11.3 & \multicolumn{1}{c|}{0-50} & 8 & 9 & \multicolumn{1}{c|}{0-46} & 10.6 & 10.3 & 0-50 \\
Medication* & 61/98 & 62.2\% & \multicolumn{1}{c|}{} & 68/169 & 40.2\% & \multicolumn{1}{c|}{} & 20/73 & 27.4\% & \multicolumn{1}{c|}{} & N/A & N/A &  \\ \hline
\end{tabular}
*Medication reported depends on the diagnostic considered. Panic disorder: Selective serotonin reuptake inhibitors (SSRIs) and Anxiolytics used in the last three months. Social anxiety disorder: Psychotropic medication in the last three months. Depression: Use of antidepressants during the treatment period (data collected post-treatment).
\label{table:demographics}
\end{table*}

All three treatments were module-based and lasted a maximum of 14 weeks. The panic disorder and social anxiety disorder G-ICBT were both comprised of 9 modules, while the depression G-ICBT was comprised of 8 modules. In all cases, the therapist provided guidance through a secure email system using an average of 10-15 minutes per week per patient. For a detailed description of each treatment, see~ \cite{eMeistring_panic_disorder} for Panic Disorder, \cite{eMeistring_social_anxiety_disorder} for Social Anxiety Disorder and \cite{depression_eMeistring} for Depression. 

\subsection{Login/logout Data as a Minimally Sensitive Source of Information from a Regulatory Perspective}

The GDPR regulates the use of personal data, such as its acquisition, processing and storage, for any entity conducting activity within the EU. Following the framework of the GDPR, the more risk the acquired data represents, the more stringent the requirements are for its processing. This risk is determined by multiple factors, including the context in which the information is collected, the sensitivity it represents and how easy it is to identify a person from it. Determining the level of risk and the resulting measures that must be taken to protect the data requires a complex legal analysis which would be outside the scope of this work. Instead, this subsection presents an initial legal argument as to why although login/logout timestamps are considered sensitive data when collected within a healthcare context, they represent the least sensitive type of information that can be extracted from an IDPT setting for user-adherence forecasting.

According to the GDPR article 4 (15) and recital 35, personal data relating to health, referred to as “health data”, may be defined as any type of information collected within the provision of healthcare services, which may reveal information relating to the current or future physical or mental health status of the person. Thus, information contained within medical records (e.g. psychometrics, medications, diagnosis, clinical self-report) are unsurprisingly understood to be defined as health data regardless of the context in which they are collected, as they in themselves would be able to reveal information relating to the health status of the person. Contrastingly, other sources of data such as demographic characteristics, login/logout timestamps, mouse and keystrokes dynamics would not in themselves reveal any information about the health status of the person but are nevertheless considered health data when collected during the provision of healthcare services such as an IDPT.  However, there exists an important distinction between these sources of information. Both mouse and keystrokes dynamics may also be categorized as biometric data (see article 4 (14)), as they through certain processing can be used to identify a person~\cite{developments_biometric_technologies, biometric_recognition_and_behavioral_detection}.

Pursuant to article 9 of the GDPR, using any source of information belonging to a special category of data, of which both health and biometric data fall under, is prohibited, unless strict requirements are met. Further, following article 32 the conditions associated with using this data (e.g. transmission, storage) must reflect the level of risk associated with processing the information. Thus, while all data collected during an IDPT would presumably be classified as belonging to a special category of data under the GDPR, ultimately, it is the level of risk they represent that will be the main bottleneck in terms of implementation. Notably, login/logout timestamps do not carry any inherent information that places it in a special category of data, rather it is categorized as such solely due to the context in which it is collected. Therefore, login/logout timestamps may be considered on the peripheral of what article 9 of the GDPR intends to prohibit and as such will be in the lower bound of the risk associated with its processing. Note that, other metrics such as number of words written, or certain demographic information might also share this characteristic. However, what makes login/logout timestamps an attractive metric is that they can be derived entirely from outside the intervention, meaning that it circumvents the need for high-risk processing (e.g. counting the number of words written from a sensitive text) to produce low-risk metrics (e.g. number of words). Finally, combining multiple low-risk sources of information may heighten the risk to the data-subject and would require increased justification due to the data minimisation principle. Thus, relying solely on login/logout timestamps for adherence forecasting within a healthcare context may substantially lower the burden of implementation and offers a promising source of information to be used in real-life applications.

\subsection{Adherence Definition}
In this work, a patient is considered adherent when achieving a \textit{sufficient} level of interactive engagement with the G-ICBT. Engagement, from a user-behavior perspective, is typically characterized through the following parameters~\cite{definition_engagement}: \textit{Amount} (i.e. length of each interaction), \textit{duration} (i.e. the period of time over which the participant is exposed to the intervention), \textit{frequency} (i.e. number of times the participant partakes in the intervention over a given period of time) and \textit{depth} (i.e. which part(s) of the intervention the participant is interacting with). The depth information however would require the data processor to have access to additional and more sensitive data and thus, the definition of sufficient engagement employed in this work is as follows:
\begin{itemize}
    \item Duration: The participant had to engage with the platform over a period lasting at least 56 days (8 weeks), meaning that they had at minimum 6-7 days per module.
    \item Frequency: The participant had to connect to the platform at least 8 times, which corresponds to the minimum number of connections required to finish the depression G-ICBT (the other two would require one more connection).  
    \item Amount: To be considered, a connection had to last more than a minute so as to exclude logins that would happen by mistake. 
\end{itemize}
Note that, in general, defining adherence through user-behavior engagement will inevitably vary depending on both the context and structure of the G-ICBT and also on what the practitioner considers sufficient engagement. It should then be expected that this variance in definition will have an impact on the machine learning model's performance, as the mapping to be learned will be different. Therefore, as to reduce the risk of positively biasing the definition towards the proposed method, the values used in this work's definition of adherence were selected prior to working with the dataset. Further, these values were selected to represent a lower-bound of sufficient engagement, as this work's hypothesis was that such a definition would be the hardest for a model to learn. Thus, the impact of raising the thresholds to consider a participant adherent on the proposed method's performance will also be explored (see Section~\ref{presentation_ablation_studies}).

\section{Adherence Forecasting}
\label{adherence_prediction_section}

Within this work, adherence forecasting was performed using between 7 to 42 days of user-interaction data (login/logout). In other words, the model had to predict at least two weeks in advance whether a patient would end up being adherent or not.
The lower-bound cutoff was selected to provide a full week-cycle of login/logout data as the participant’s weekly schedule might strongly influence their pattern of interaction with the treatment. An upper-bound cutoff at day 42 was selected as some symptom trajectories were shown to experience more changes around week 5-6~\cite{trajectory_symptoms_change_small_proportion_at_week_5, trajectory_symptoms_can_have_changepoint_at_the_6_weeks_mark}.

To avoid indirect overfitting of the dataset when designing the model, the participants were divided into two non-overlapping subsets. The first subset referred to as the \textit{Exploration Dataset}, contained the data from 100 randomly selected participants. The exploration dataset was employed for feature exploration, architecture building and hyperparameter optimization. The second subset referred to as the \textit{Main Dataset}, contained the remaining 242 participants. Cross-validation with a fold size of 10 using the main dataset was employed to evaluate the selected model from the Exploration Dataset. Note that when training a model to be evaluated, the Exploration Dataset was concatenated to the training fold but was not used as part of the testing folds. For reproducibility purposes, the participants' indexes associated with the exploration dataset, which were randomly selected, are made available on this work's GitHub. 

\subsection{Data Pre-Processing}
Considering the first login data entry as the start of the first session of the G-ICBT, the following features were computed each day from the login/logout data: 
\begin{itemize}
    \item Whether or not the participant logged in during this particular day (0 or 1). 
    \item The total time the participant was logged during this particular day.
\end{itemize}

Then, feature-wise scaling was performed such that the values were centred around the mean with a unit standard deviation:

\begin{equation}
    \mathbf{X}' = \frac{\mathbf{X}-\mu}{\sigma}
\end{equation}
Where $\mu$ and $\sigma$ represent the mean and standard deviation of the current training set fold.

Then, for each participant, these feature vectors were aggregated sequentially to form a first $2\times7$ matrix that represented the first 7 days of a given user's interaction data. A copy of this initial matrix was then created and concatenated with the feature vector of day 8 to form a new matrix. This process was then repeated for day 9, 10 and so on. Thus, a total of 35 matrices (examples) with a number of columns ranging from 7 to 42 were created for each patient contained within the dataset. This sequential data representation had two objectives: 1) Reducing the potential loss of information from aggregating the login/logout data to form a single feature vector. 2) Enabling the same trained model to forecast the adherence of a user at any point during the considered time span (7 to 42 days).

\subsection{Self-Attention Network}
\label{SelfAttentionNetworkDescription}
\begin{figure*}[!htbp]
\centering
\includegraphics[width=\linewidth]{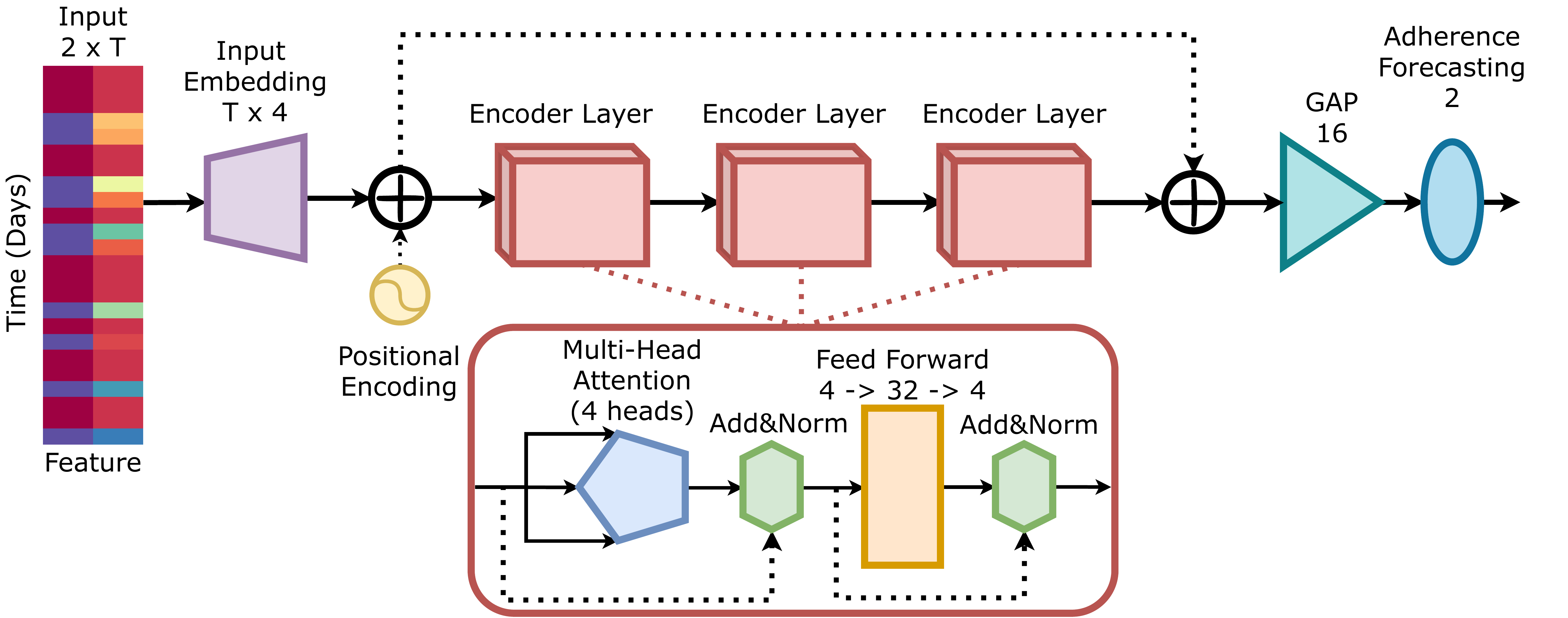}
\caption{The Self Attention Network's architecture employed for adherence forecasting contains 1186 learnable parameters. GAP refers to the Global Average Pooling operation. The plus sign refers to element-wise summation. A dropout of 0.1 is also applied in the Multi-Head Attention (MHA) module, immediately after the MHA and in the Feed Forward module. Note that the input fed to the network is of shape Tx2 (shown transposed in the figure), where T represents the length (in days) of the example, which is variable (between 7 and 42 days in this work).} 
\label{fig:self_attention_network}
\end{figure*}

A compact Self-Attention-based network inspired by~\cite{transformer} was designed to leverage the sequential information that naturally arises within the data generated from this work's context. The network's architecture is presented in Figure~\ref{fig:self_attention_network}. AdamW~\cite{adamW} was employed for the network's optimization with a batch size of 64. The variable sequences' lengths were padded when creating each batch and a mask of the padding was used when feeding the batches to the network. The learning rate (lr=0.001306) was selected from the Exploration Dataset by random search~\cite{random_search} using a uniform random distribution on a logarithmic scale between $10^-5$ and $10^0$ with 100 candidates. During the random search, the following hyperparameters were also considered: 
\begin{itemize}
    \item Input embedding size [1, 2, 4, 8, 16, 32, 64, 128]. Value selected: 4
    \item Number of heads: [1, 2, 4, 8]. Value selected: 4
    \item Number of neurons in the feed forward network's hidden layer: [1, 2, 4, 8, 16, 32, 64, 128]. Value selected: 32
    \item Dropout: [0., 0.1, 0.2, 0.3, 0.4, 0.5]. Value selected: 0.1
    \item Number of Encoder Layer: [1, 2, 3]. Value selected: 3
\end{itemize}

10\% of the training data was held out as a validation set to perform early stopping (20 epochs threshold). Additionally, learning rate annealing with a factor of five and a patience of ten was also employed. Finally, to alleviate the effect of using an imbalanced dataset ($\sim$30\% of participants were non-adherent to the treatment), a per-class weighted cross-entropy loss was employed as the criterion to optimize during training where the per-class weight was obtained as follows:
\begin{equation}
    \frac{1}{\text{\# of examples from the given class}}
\end{equation}
In other words, the network was punished further when making mistakes on the underrepresented class (dropout) than on the overrepresented class (adherent) proportionally to each class' size.

In this work, the previously described trained network will be referred to as the \textit{Self-Attention Network}. The Self-Attention Network's implementation written with PyTorch~\cite{PyTorch} and training procedure are made readily available here: \href{https://github.com/UlysseCoteAllard/AdherenceForecastingG-ICBT}{github.com/UlysseCoteAllard/AdherenceForecastingG-ICBTl}.

\subsection{Threshold for Success}
Following the practice established in~\cite{predicting_treatment_failure_gICBT}, this work considers three empirical thresholds to contextualize the classifier's performance. Importantly, when considering these thresholds one should look not only at if they are exceeded, but also how early in the treatment the model's forecast can surpass them. Note that all thresholds are defined using the \textit{balanced accuracy} which is defined as follows: 
\begin{equation}
    \text{Balanced Accuracy}=\frac{1}{2}\left( \frac{TP}{TP + FN} + \frac{TN}{TN + FP} \right)
\end{equation}

where TP, TN, FP and FN stands for true positive, true negative, false positive and false negative respectively. Balanced accuracy was employed over the simple accuracy in this work to avoid inflating the performance estimates due to the class-imbalance present in the dataset.

\subsubsection{Threshold 1: Better than random}
The first threshold assesses whether or not the model's forecasting of user-adherence can perform better than chance. This was evaluated by considering if the lower bound of 95\% confidence interval for the balanced accuracy was above 50\%. 

\subsubsection{Threshold 2: Minimal threshold where clinicians will take action}
Eisenberg and Hershey~\cite{threshold_test_treatment_65_to_70_percent} reported that clinicians were willing to take action based on predictions once their accuracies reached 65-70\%. As the main purpose of an adherence forecasting system for G-ICBT is to empower the clinician to perform more precise targeted interventions, whether they would be willing to act on the system's prediction is critical. Thus, the second threshold was considered to be reached when the lower bound of the 95\% confidence interval for the balanced accuracy surpassed 65\%. 

\subsubsection{Threshold 3: Threshold where clinicians will take action}
The third threshold was considered to be reached when the lower bound of the 95\% confidence interval for the balanced accuracy surpassed 70\%. 

Note that while threshold 2\&3 were defined based on preliminary work~\cite{threshold_test_treatment_65_to_70_percent} that is nowadays relatively outdated, said work was to the best of our and previous work's knowledge~\cite{predicting_treatment_failure_gICBT} the only empirical data currently available. 

\subsection{Ablation Studies}
\label{presentation_ablation_studies}
Three ablation studies were conducted to quantify the impact of some of the decisions made when designing the model.
\subsubsection{Weighting of the loss function}
A first ablation study was performed to evaluate the importance of the per-class weighting. The comparison was made by training the network with and without weighting each example based on their associated class prevalence. In both cases, the average balanced accuracy was computed over 20 independent runs.

\subsubsection{Training with a fixed sequence length vs all sequence lengths simultaneously}
To better understand the effect of training the network with multiple sequence lengths simultaneously, a second ablation study was conducted. The performance of the Self-Attention Network when training over multiple sequence lengths was compared to training the network using only information from a single, fixed sequence length. This comparison was made using the average balanced accuracy computed over 20 independent runs for both methods. The sequence lengths selected were based on the days that the model surpassed threshold 1, 2 and 3 as well as the longest sequence length considered in this work (42 days). Note that for comparison fairness, hyperparameter optimization using random search, as described in Section~\ref{SelfAttentionNetworkDescription} was performed for each fixed sequence length considered.

Mann-Whitney-U~\cite{mann_whitney_u}, a non-parametric null-hypothesis significance test for unpaired data, was applied to compare if training the classifier with examples aggregated from different sequence lengths led to a different performance than training the same model with fixed-sequence length.

\subsubsection{Alternative definition of adherence}
As previously stated, what is considered \textit{sufficient engagement} from the participants varies based on both the context and structure of the G-ICBT. Therefore, to evaluate the ability of the proposed approach to cope with different and more stringent definitions of adherence, two alternatives adherence definition were proposed. Table~\ref{table:alternative_definition_adherence} summarizes the two alternative definitions and compares them with the one used throughout this work. Note that these alternative definitions were defined solely to evaluate the adaptability of the proposed approach and as such they do not have an explicit clinical basis. For comparison fairness, hyperparameter optimization using random search, as described in Section~\ref{SelfAttentionNetworkDescription} was performed for each new adherence definition.

\begin{table}[!htbp]
\centering

\caption{The two alternative definitions of adherence compared to the original definition.}
\begin{tabular}{@{}cccc@{}}
\toprule
 & \textbf{Original} & \textbf{Alternative A} & \textbf{Alternative B} \\ \midrule
\textbf{Duration} & Min. 56 days & Min. 56 days & Min. 56 days \\
\textbf{Frequency} & 8 connections & 12 connections & 16 connections \\
\textbf{Amount} & 60 seconds & 150 seconds & 300 seconds \\
\multicolumn{1}{l}{\textbf{\begin{tabular}[c]{@{}l@{}}\% Participants\\  Non-Adherent\end{tabular}}} & $\sim$30\% & $\sim$49\% & $\sim$74\% \\ \bottomrule
\end{tabular}
\\*The number of connections stated in \textbf{Frequency} have to occur over at least 56 days. Additionally, each connection has to last longer than the number of seconds stated in the \textbf{Amount} row, to be considered. 
\label{table:alternative_definition_adherence}
\end{table}

\section{Results}
\label{results_section}
In this section, all evaluation metrics were computed using the scikit-learn python library version 1.0.2~\cite{scikit_learn}.

\subsection{Results of Adherence Forecasting}
\begin{figure*}[!htbp]
\centering
\includegraphics[width=\linewidth]{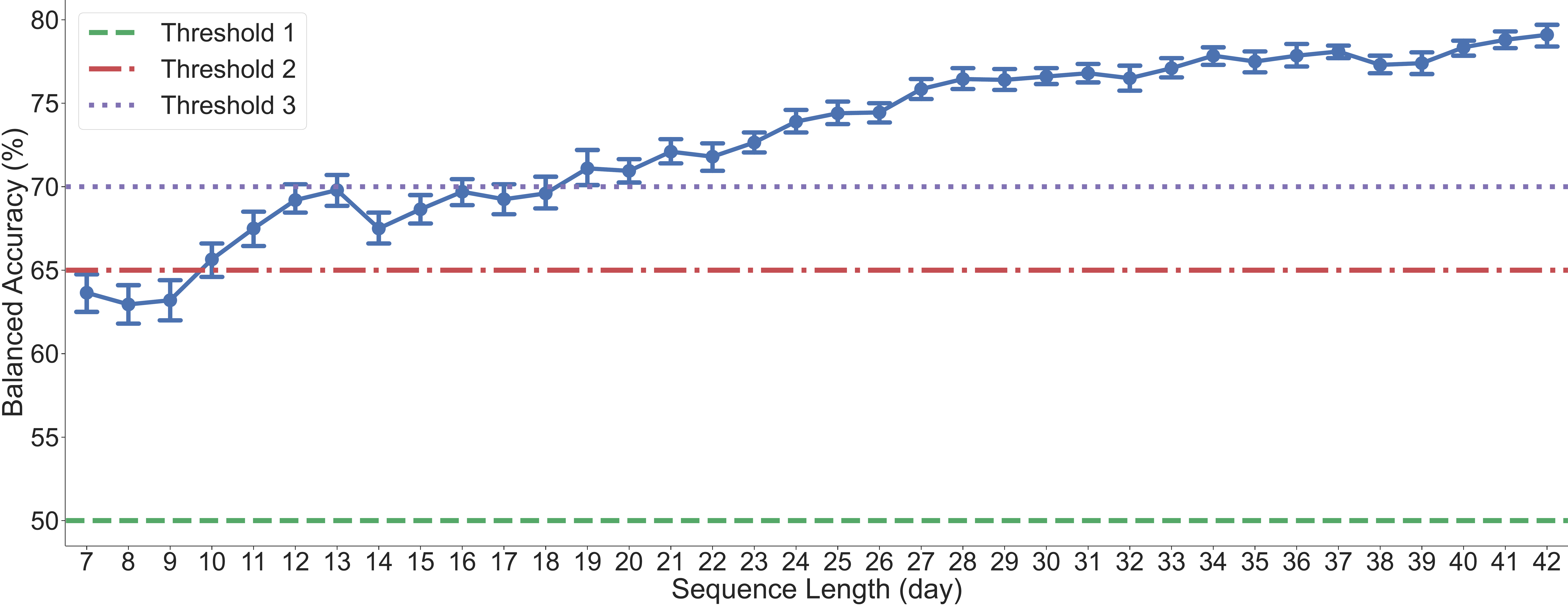}
\caption{Point plot of the average balanced accuracy over 20 runs with respect to the number of days considered when forecasting the user's adherence. The error bars represent the 95\% confidence interval of the balanced accuracy. The first threshold (green dashed line) is exceeded on day 7 (smallest sequence length considered), the second (red dash-dotted line) on day 11 and the third (purple dotted line) on day 20.}
\label{fig:results_self_attention_multi_sequence}
\end{figure*}

Figure~\ref{fig:results_self_attention_multi_sequence} shows the average balanced accuracy over time computed over 20 independent runs. The first threshold is exceeded on day 7 (corresponding to the smallest sequence length considered by the model). The second and third thresholds are surpassed on day 11 and 20 respectively.   

\begin{figure}
    \centering
    \includegraphics[width=\linewidth]{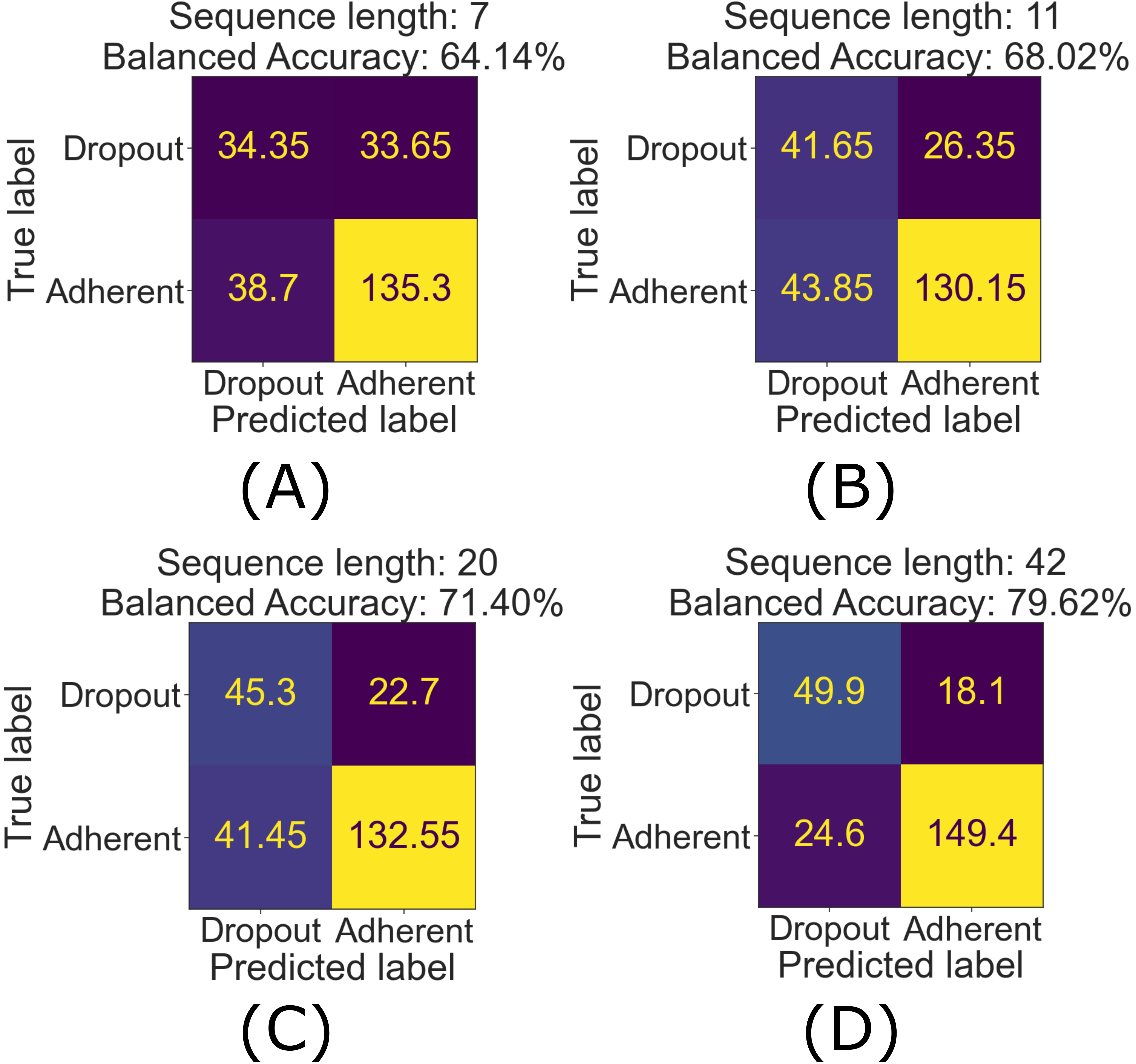}
    \caption{Confusion matrices representing the average number of participants being classified into each class by the Self-Attention Network over 20 independent runs. (A), (B) and (C) showcase the confusion matrices for the first time the network manages to surpass one of the defined thresholds for success (on day 7, 11 and 20 for threshold 1, 2 and 3 respectively). (D) shows the confusion matrix for the sequence length of 42 days (the upper bound).}
    \label{fig:confusion_matrices}
\end{figure}

\begin{figure}
    \centering
    \includegraphics[width=\linewidth]{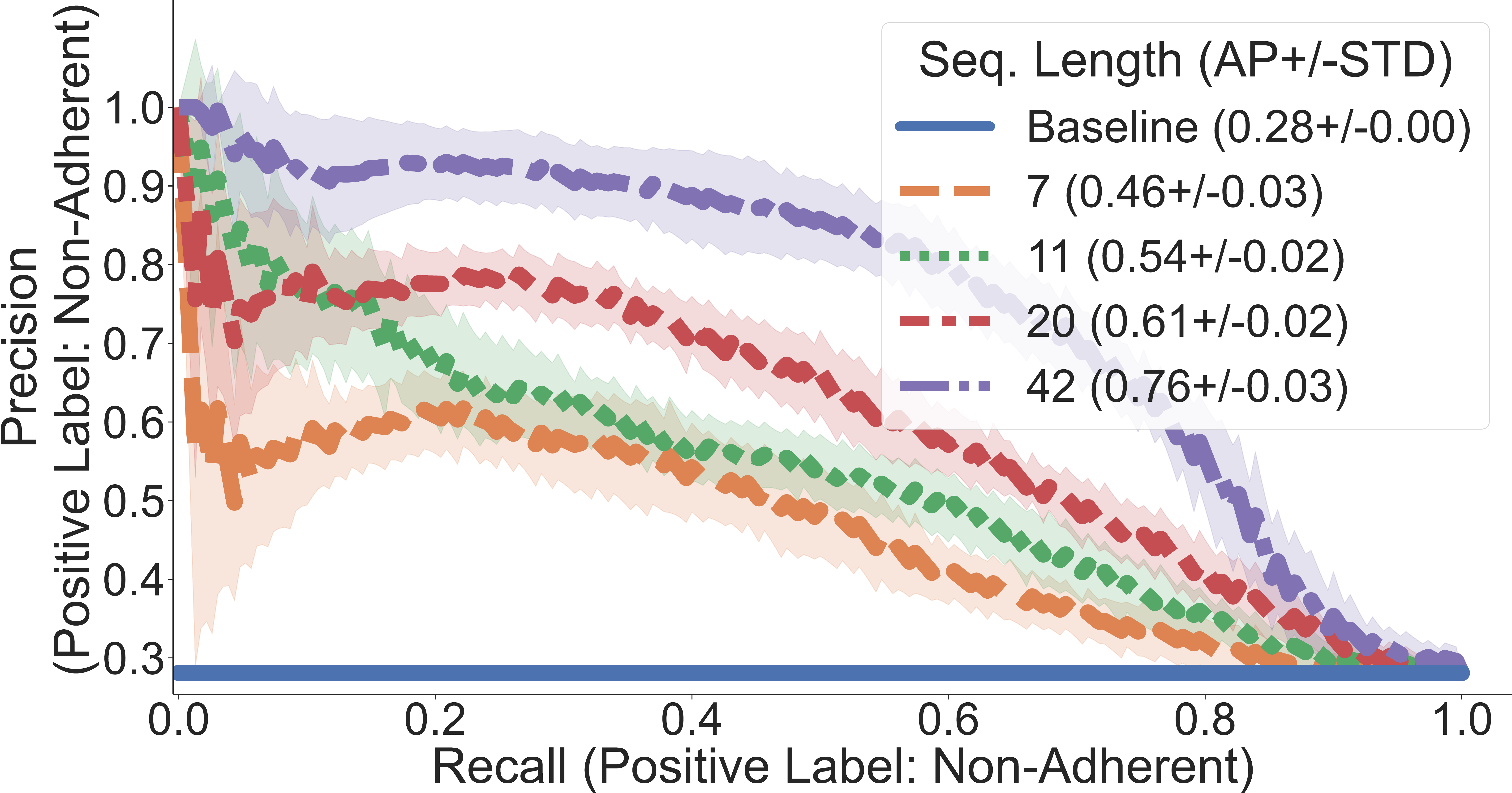}
    \caption{Precision-Recall Curves and their associated Average Precision (higher is better) over 20 runs for sequence lengths of 7, 11, 20 and 42 computed over 20 independent runs. The blue line (Baseline) shows the expected performance of a random classifier corresponding to the proportion of non-adherent labels. The shaded areas represents the standard deviations. AP stands for Average Precision which is the estimator for the Precision-Recall Area Under the Curve (PR-AUC). STD for Standard Deviation.}
    \label{fig:pr_curve}
\end{figure}

To provide a more holistic view of the Self-Attention Network's performance, Figure~\ref{fig:confusion_matrices} presents the confusion matrices for day 7, 11, 20 (the threshold-surpassing day) and 42 (the upper bound sequence length). The values reported in these confusion matrices correspond to the average number of true positive, true negative, false positive and false negative as predicted by the Self-Attention Network over 20 independent runs. Additionally, Figure~\ref{fig:pr_curve} shows the Precision-Recall Curve and Precision-Recall Area Under the Curve (PR-AUC) of the network's performance for these same four sequence lengths. Note that the PR-AUC is calculated using the Average Precision as suggested in~\cite{pr_curve_and_average_precision}. These metrics were computed using the Scikit-Learn V1.0.2~\cite{scikit_learn}

\subsection{Ablation Studies}
\label{AblationStudies}
\subsubsection{Weighting of the loss function}
\begin{figure}[!htbp]
\centering
\includegraphics[width=\linewidth]{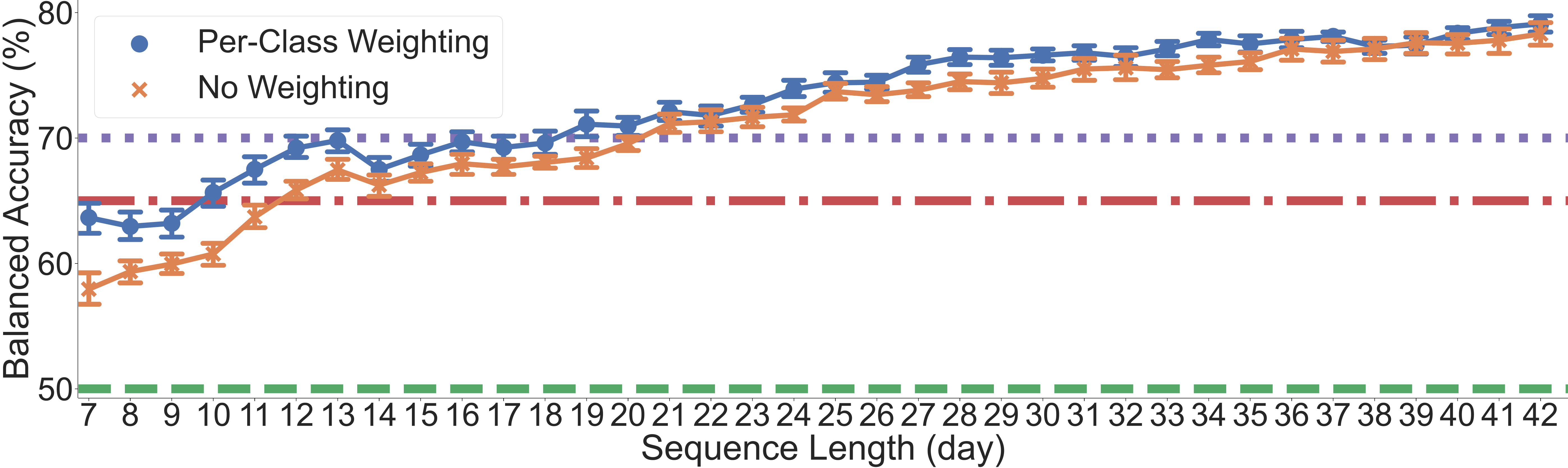}
\caption{Point plot comparison of the average balanced accuracy between training the Self Attention Network with and without per-class weighting over 20 runs. The blue dotted line corresponds to the network trained with per-class weighting, while the full orange line corresponds to the network trained without. The error bars correspond to the 95\% confidence interval of the balanced accuracy. The three thresholds for success considered in this work are shown as horizontal lines to help contextualize the performance of both training approaches.}
\label{fig:weigthing_results_comparison}
\end{figure}

Figure~\ref{fig:weigthing_results_comparison} presents the average balanced accuracy using the Self-Attention Network trained with and without per-class weighting. 

\subsubsection{Training with a fixed sequence length vs all sequence lengths simultaneously}
\begin{figure}[!htbp]
\centering
\includegraphics[width=\linewidth]{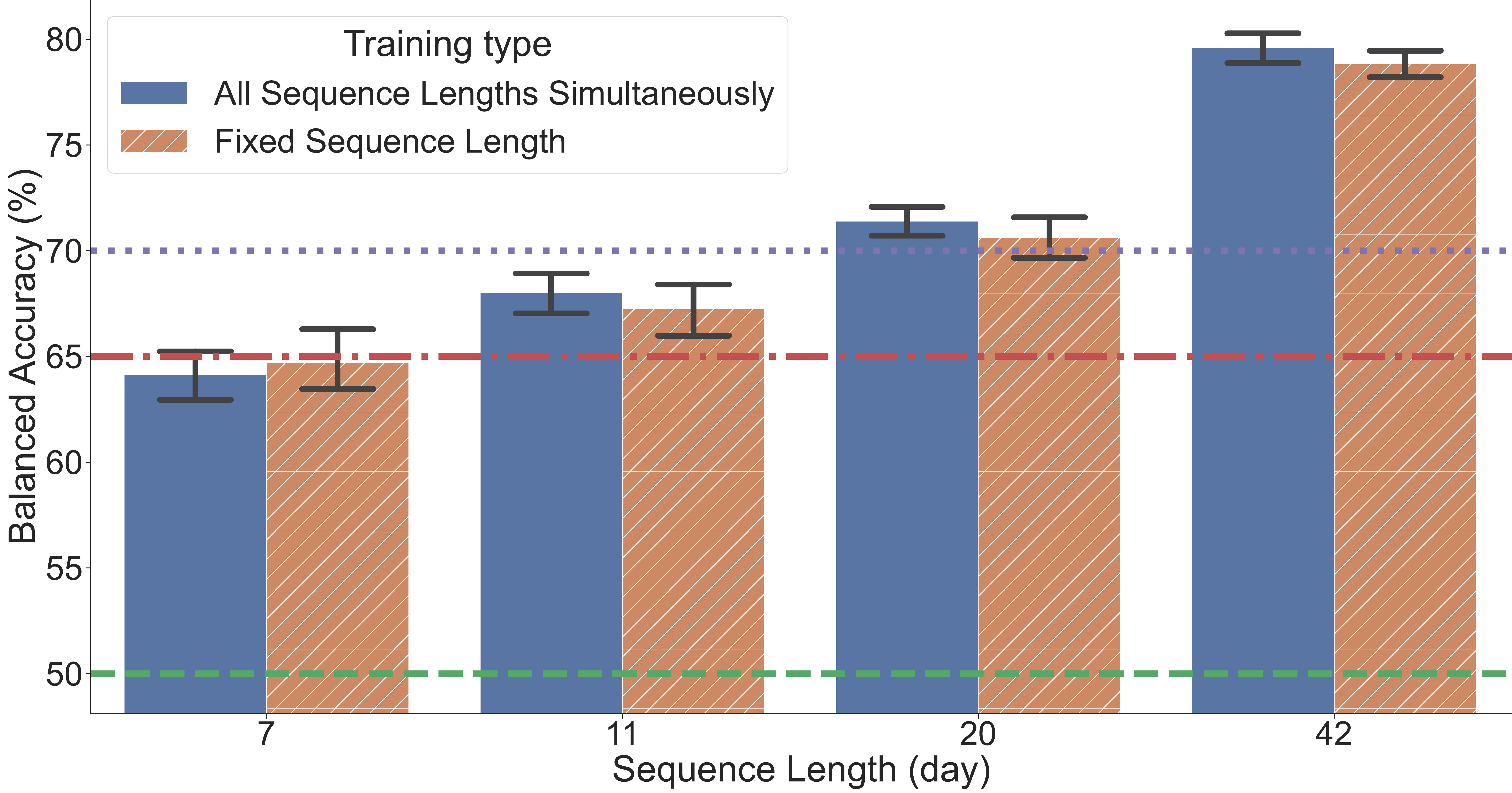}
\caption{Comparisons of the average balanced accuracy over 20 runs when training the Self-Attention Network over multiple sequence lengths simultaneously versus training the model using only data from a single sequence length. The Self Attention Network trained over multiple days is represented by the blue bar plots. The networks trained over a single day are represented by the hashed orange bar plots. The error bars represent the 95\% confidence interval of the balanced accuracy. In the case of single day training, random search for hyperparameter selection is performed independently for each sequence length. The horizontal lines represent the three thresholds for success considered.}
\label{fig:multi_sequence_training_vs_single_sequence_comparison}
\end{figure}

Figure ~\ref{fig:multi_sequence_training_vs_single_sequence_comparison} presents the performance of the Self-Attention Network when trained over multiple sequence lengths. Note that for comparison fairness, hyperparameter optimization using the random search described in Section~\ref{SelfAttentionNetworkDescription} is performed for each fixed sequence length considered.  

For all sequence lengths (7, 11, 20 and 42 days), the null hypothesis cannot be rejected ($p>0.05$) according to the Mann-Whitney-U test (see Table~\ref{table:U_test_details} for details). Thus, there is no statistical significant difference between the two methods.

\begin{table}[!htbp]
\caption{Results of the Mann-Whitney-U tests on the models trained with a fixed sequence length versus models trained with all sequence lengths.}
\centering
\begin{tabular}{@{}ccccc@{}}
\toprule
 & \multicolumn{4}{c}{\textbf{Sequence Length}} \\ \midrule
 & \textbf{7} & \textbf{11} & \textbf{20} & \textbf{42} \\
\textbf{U-Value} & 198.0 & 152.0 & 156.5 & 139.0 \\
\textbf{p-Value} & 0.9681 & 0.2005 & 0.2460 & 0.1010 \\ \bottomrule
\end{tabular}
\\[2pt]
*The null hypothesis that the two populations are equal is rejected at p$<$0.05.
\label{table:U_test_details}
\end{table}

\subsubsection{Alternative definition of adherence}
\begin{figure}
    \centering
    \includegraphics[width=\linewidth]{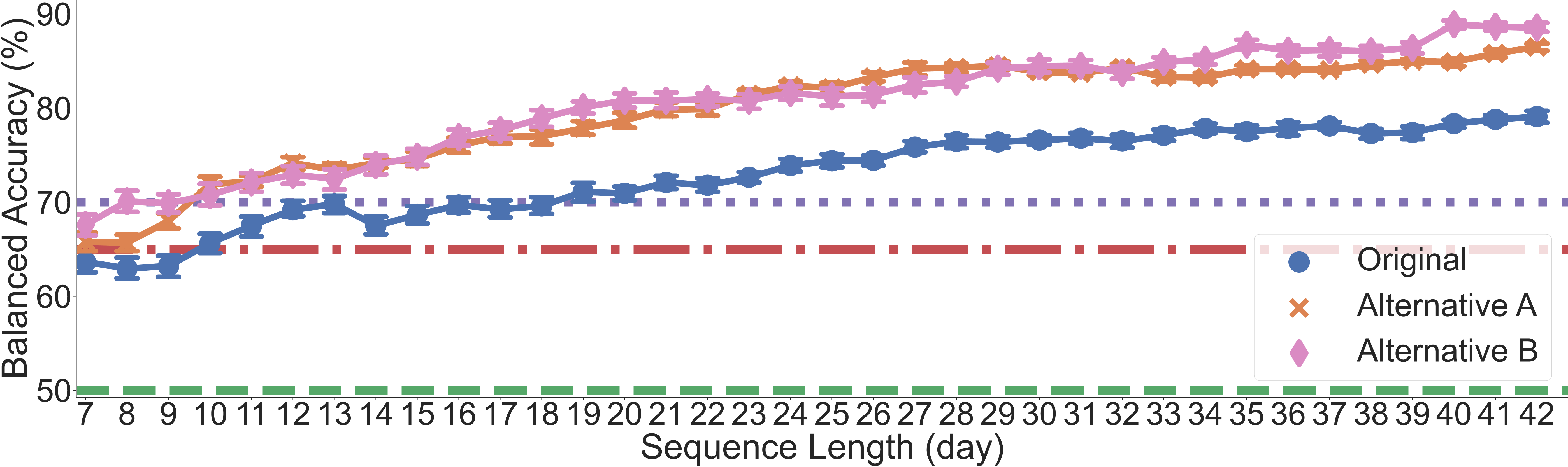}    \caption{Point plot comparison of the average balanced accuracy over 20 runs when predicting different levels of user engagement. Original (the blue line) corresponds to the definition of adherence used in this work. Alternatives A and B correspond to more stringent definitions of user engagement as defined in Table~\ref{table:alternative_definition_adherence}. The error bars correspond to the 95\% confidence interval of the balanced accuracy.}
    \label{fig:adherence_definition}
\end{figure}

Figure~\ref{fig:adherence_definition} shows a point plot of the model's performance based on the different adherence definitions considered.

\section{Discussion}
\label{discussion_section}
Achieving robust adherence forecasting within the context of G-ICBT would enable better management of the limited resources available within mental healthcare. In practice, such capabilities would help clinicians identify which patients are more likely to dropout, fostering more targeted intervention to increase retention rates or allowing earlier redirection of individuals towards more suitable treatment. A substantial hindrance for such a goal to be achieved however, is the validation and deployment of adherence forecasting models, as they generally require highly sensitive data from individual patients for accurate forecasting to occur~\cite{wallert2018}. This highly sensitive data problem might be one of the main reasons why adherence prediction for G-ICBT using ML is still understudied. As such, this paper proposes performing adherence forecasting on an individual basis using ML by solely leveraging non-sensitive information. However, with a decrease in data-sensitivity comes a decrease in the quantity and quality of information that may be derived from patients during their G-ICBT treatment.

The signal that can be derived from login/logout data is inherently noisy as one cannot distinguish a user's meaningful interaction with the program from an idle connection to the service. Thus, the major finding in this study is that treatment adherence to G-ICBT can be predicted with a balanced accuracy above 70\% after 20 days (when only 36\% of the treatment is completed) despite relying solely on login/logout data. This well exceeds the performance of a random classifier and is in line with clinicians' subjective preference of "good enough" predictive ability~\cite{threshold_test_treatment_65_to_70_percent}. This finding is reinforced by Figure~\ref{fig:confusion_matrices} and \ref{fig:pr_curve} as it can be seen that the proposed model is relatively robust to the class imbalance and overall tends to be more pessimistic regarding the likelihood of a patient to be adherent to the treatment.

Figure~\ref{fig:weigthing_results_comparison} shows that using per-class weighting systematically improves the forecasting performance at the beginning of the treatment and overall consistently performs equal to or better than not using per-class weighting. Thus, considering that this type of weighting comes at no extra cost during both the training and inference phase, its use should be considered a net benefit to the model. 
Interestingly, from Figure~\ref{fig:multi_sequence_training_vs_single_sequence_comparison} and from the Mann-Whitney-U test, it can be seen that having a single model able to predict multiple sequence lengths does not hinder performance compared to having a different model for each specific sequence length. A possible explanation is that the relatively small dataset used could have advantaged the multi-sequences approach, as considering sequence lengths of various sizes from the same patient acted as a form of data augmentation. Nevertheless, given this work's context, small datasets (when compared to other areas in ML) is a reality that has to be contended with, at least in the near future. Further, for real-life implementation, it is desirable that a single model is able to contend with any sequence length as otherwise, one would have to deploy one model per time-step considered, onto the G-ICBT platform, thus rapidly increasing maintenance costs. Such considerations also naturally hold for ML algorithms unable to contend with temporal data without relying on feature extraction to characterize the signal as a pre-processing step (e.g. support vector machine~\cite{svm}, random forest~\cite{random_forest}, linear discriminant analysis~\cite{LDA}). Interestingly, the results presented in Figure~\ref{fig:adherence_definition} support the initial hypothesis that the proposed model will, in general, more easily forecast adherence when its definition is based on stricter requirements for user-engagement. This characteristic in conjunction with the model's ability to easily adapt to new adherence definitions is important to enable the clinician to nuance the level of user-engagement required, to help underpin their decision making in relation to the patient adherence.

Beyond login/logout timestamps, other types of data that also appear minimally sensitive could be useful in predicting patient adherence in Internet-delivered Cognitive Behavioral Therapy (ICBT). For example, in a trial of ICBT for patients who reported symptoms of anxiety, depression or both following a myocardial infarction event, Wallert et al.~\cite{wallert2018} showed that the number of words written in the first homework was the third strongest predictor of adherence, after self-assessed cardiac-related fear and the patient's sex. Thus, the number of words written during the first interaction (or similar statistics) with the online platform could represent an interesting possibility for further investigation in minimally data-sensitive approaches to adherence forecasting. However, two important caveats have to be made. 1) Because these statistics would be derived from highly sensitive information (the patient's thoughts within the context of a medical intervention), the data controller (e.g. the hospital) would need to process the data and derive the relevant minimally sensitive statistics. Otherwise, the sensitive data would have to be sent to the data processor (e.g. the company providing the adherence prediction) and thus the more stringent regulations would automatically apply, defeating the purpose of a minimally data-sensitive approach. Contrastingly, an advantage of using login/logout timestamps is that they are not derived from sensitive data and as such can be transmitted to the data processor as-is. 2) As previously stated, one of the cornerstones of the GDPR and similar regulations is data-minimization. As such, the amalgamation of data (e.g. using login/logout timestamps in conjunction with the number of words written in the first homework) to perform adherence prediction, needs to improve the performance of the system to a degree that would justify the corresponding increase in collected data. Notably, this is especially pertinent as this work has shown that login/logout timestamps, on their own, provide information that can be leveraged to perform adherence forecasting for patients undergoing G-ICBT treatment. Therefore, login/logout data intrinsically contains information that can be used to predict the behavior of an identifiable natural person (as defined within the GDPR). In other words, from a regulatory perspective login/logout timestamps are no longer subject to the requirements in the GDPR solely due to them being collected in a healthcare context, but instead due to the information the data contains as has been illustrated in this work. Consequently, following the core principle of data minimization and in light of this work, it seems that even stronger justification would be needed to use additional and/or more sensitive data to perform such forecasting.


Automatic adherence forecasting for G-ICBT should be viewed as a tool that can supplement available mechanisms used to follow patients during their treatment. Relying solely on the prediction of these models to identify potentially non-adherent users poses the risk of masking certain categories of patients which, for example, might not have been sufficiently represented in the training dataset. 
Further, an important challenge that remains to be addressed for the applicability of automatic adherence forecasting in a real-world context is the feedback-loop effect that will result from the clinician considering the model's inference when deciding which patients require a more targeted intervention. In other words, it should be expected that the model predicting: "Patient A: Non-Adherent", might result in the patient factually finishing the program (through the clinician acting on this prediction), even if the counterfactual (i.e. if the model was not used) would have resulted in the patient dropping out. This challenge will be especially critical when considering how to update the adherence forecasting model over time as data collected while the system is active will necessarily be biased with this model-clinician-patient interaction.

Interestingly, if a tool using the proposed approach to automatically forecast patient’s treatment adherence came to be used in practice, it would highlight and possibly valorize login/logout timestamps as a meaningful source of information for the guiding psychologist. As this information can easily be accessed by the clinician, it is possible that they would, in time, learn to predict the patient’s treatment adherence via these data. Thus, beyond the legal motivation of using a minimally-data sensitive approach, relying on a small amount of interpretable information might have the added effect of fostering this type of learned association from the guiding psychologist. Evaluating whether this potential learned mapping would be useful or detrimental to the interactions with the patients (both due to the feedback loop previously mentioned and the fact that the learned model itself will not be perfect) is however, outside of the scope of this work. Nonetheless, this is another factor that will have to be considered for the eventual deployment of such a tool on a G-ICBT platform.

\subsection{Limitations}
This work's main limitation is also its main motivation. Being able to share and utilize data which originates from ongoing patient treatment to benchmark new ML models is understandably highly restricted and regulated. As such, for the dataset used in this work to be made available as a future benchmark, substantial information removal and pseudonymization had to be performed. Thus all information regarding demographic, clinical (including followed treatment) and interaction data (excluding login/logout) from the patients had to be stripped away from the dataset used in this work. 

As a direct consequence of the strict information removal, the proposed algorithm could not be compared against a model having access to more meaningful information (e.g. demographic, psychometric, linguistic) from the patients, to contextualize the impact of using the proposed minimally-invasive approach. 

Another limitation was the absence of a gold standard to benchmark the proposed approach against. This limitation was an additional motivation for making the dataset used in this work public, so that new methods can be compared to each other more easily.

It is also important to highlight that to obtain a sufficiently large dataset, the three populations of patients having a primary diagnosis of panic disorder, social anxiety disorder or depression were aggregated together.
However, due to the pseudonymization processing which took place prior to this work, the patient's diagnostic was not available within the dataset used. Unfortunately, as a consequence of the nature, symptomatology and course of disease of these disorders, there may exist differences regarding both patients' adherence and adherence forecasting accuracy between these three populations. Thus, while this work showed that it is possible to perform automatic adherence forecasting based solely on login/logout timestamps generated from a G-ICBT treatment, future works will investigate how these performances vary across primary diagnosis and G-ICBTs.

\section{Conclusion}
This paper presents a minimally data-sensitive approach, based on a self-attention network, to perform adherence forecasting of patients undergoing G-ICBT. Overall, the proposed approach was shown to reach an average balanced accuracy above 65\% with a confidence of 95\% on day 11 ($\sim$20\% of the treatment's total length) and above 70\% with a confidence of 95\% after only 20 days ($\sim$36\% of the treatments total length). Thus, the results show that login/logout information is sufficient to achieve robust adherence forecasting. Within a clinical setting, such an adherence forecasting tool could be used by the clinician to perform more targeted intervention with patients that are at risk of dropping-out. Further, because of the minimally data-sensitive approach, the additional requirements due to the context of the data, are minimized. Lastly, a core tenant of the GDPR and similar regulations is the data minimization principle. This principle refers to that only data that is strictly necessary to achieve the purpose of the data collection, shall be collected and processed. By illustrating that low-risk, minimally sensitive data such as pseudonymized login/logout data can achieve practical and useful results, a new benchmark for the data required for adherence forecasting may have been set.

Future works will focus on deploying the proposed solution within a G-ICBT platform to evaluate its usefulness within real-world applications.

\section{Acknowledgement}
This work was partially supported by the Research Council of Norway as a part of the INTROMAT project (grant agreement 259293). We would also like to thank the anonymous reviewers for their thoughtful comments and suggestions.

\bibliographystyle{IEEEtran}
\bibliography{main}

\pagebreak

\end{document}